\documentclass[runningheads]{llncs}
\usepackage{graphicx}
\usepackage{float}
\usepackage[ruled,vlined]{algorithm2e}
\usepackage{multirow}

\begin{document}
\title{To augment or not to augment? Data augmentation in user identification based on motion sensors}
\author{Cezara Benegui
		\and
        Radu Tudor Ionescu\vspace{-0.2cm}}

\institute{%Faculty of Mathematics and Computer Science\\
University of Bucharest, 14 Academiei, Bucharest, Romania\\
\email{cezara.benegui@fmi.unibuc.ro, raducu.ionescu@gmail.com}
}

\titlerunning{Data augmentation in user identification based on motion sensors}
\maketitle              % typeset the header of the contribution
\begin{abstract}
Nowadays, commonly-used authentication systems for mobile device users, e.g. password checking, face recognition or fingerprint scanning, are susceptible to various kinds of attacks. In order to prevent some of the possible attacks, these explicit authentication systems can be enhanced by considering a two-factor authentication scheme, in which the second factor is an implicit authentication system based on analyzing motion sensor data captured by accelerometers or gyroscopes. In order to avoid any additional burdens to the user, the registration process of the implicit authentication system must be performed quickly, i.e. the number of data samples collected from the user is typically small. In the context of designing a machine learning model for implicit user authentication based on motion signals, data augmentation can play an important role. In this paper, we study several data augmentation techniques in the quest of finding useful augmentation methods for motion sensor data. We propose a set of four research questions related to data augmentation in the context of few-shot user identification based on motion sensor signals. We conduct experiments on a benchmark data set, using two deep learning architectures, convolutional neural networks and Long Short-Term Memory networks, showing which and when data augmentation methods bring accuracy improvements. Interestingly, we find that data augmentation is not very helpful, most likely because the signal patterns useful to discriminate users are too sensitive to the transformations brought by certain data augmentation techniques. This result is somewhat contradictory to the common belief that data augmentation is expected to increase the accuracy of machine learning models.

\end{abstract}

\section{Introduction}

Nowadays, mobile devices have become the most utilized digital devices in our daily activities, replacing personal computers. Usage of our personal devices and access to all applications require strong authentication systems. Albeit all mobile operating systems grant users the possibility to set up secure passwords, PINs or unlock patterns, it is well known that such protection mechanisms are not fully secure and are prone to physical attacks such as fingerprint attacks \cite{Andriotis-WiSec-2013,Zhang-SPSM-2012}, or security breaches caused by internal audio or video signal hacking \cite{Simon-SPSM-2013}. A potential solution to avoid such attacks is to rely on an additional implicit authentication system. Some recent works \cite{Benegui-Access-2020,Buriro-ISBA-2017,Neverova-Access-2016,Sitova-TIFS-2016,Sun-ECML-2017} proposed such unobtrusive authentication systems based on analyzing data captured by motion sensors, e.g. accelerometer and gyroscope, using machine learning methods. 

In a realistic setting, in which implicit authentication factors (based on motion sensors) support explicit authentication factors (based on face recognition or fingerprint scanning), the registration process necessary for the implicit authentication system is expected to be short, i.e. the number of samples collected during registration must be reduced to a bare minimum. This requirement is imposed by the fact that explicit authentication systems are typically based on a fast registration process. Hence, an implicit authentication system should not represent an additional burden to the end user. In this context, machine learning models based on motion sensors should deliver good performance results in a few-shot learning context, as also noticed by Benegui et al. \cite{Benegui-Access-2020}.

In this paper, we aim to find out if the accuracy of few-shot learning models based on motion sensors can be improved through data augmentation. Although data augmentation is a commonly-used approach to enhance image \cite{Perez-Arxiv-2017,Shorten-JBD-2019} and signal \cite{Le-Guennec-AALTD-2016} processing systems, to our knowledge, we are the first to study data augmentation techniques for user authentication based on signals collected from motion sensors. We note that we cannot trivially borrow data augmentation techniques from computer vision. For example, in computer vision, flipping an image horizontally will contain the same objects in a different yet realistic pose, which should help the machine learning system to generalize better. In motion signal analysis, flipping a signal on the temporal axis will invert any patterns that belong to a user, thus having a negative effect on the machine learning model. Even the addition of random noise might be problematic, since the signal patterns specific to a user can be very sensitive and the added noise might simply cover them. In this work, we propose a set of data augmentation methods for motion signals, that represent more plausible ways of improving the generalization capacity of machine learning models for motion sensor data. In order to make sure that the original signals are not affected by excessively strong augmentation, we empirically experiment with parameters that control the degree of augmentation, finding optimal values for these parameters. Besides trying out alternative approaches to augmented data, we also seek to identify if a mixture of multiple data augmentation methods can provide better results.  
In summary, our aim is to find answers to the following research questions (RQs) in the context of few-shot user identification based on motion sensor signals:
\begin{itemize}
%\vspace{-0.2cm}
    \item RQ1: Can data augmentation bring accuracy improvements?
    \item RQ2: Which of the proposed data augmentation methods brings accuracy improvements?
    \item RQ3: Are the data augmentation methods generic or specific to certain machine learning models?
    \item RQ4: Can mixtures of data augmentation techniques bring accuracy improvements?
\end{itemize}

We hereby note that RQ1 and RQ2 are strongly related, although RQ1 is more generic. If at least one of the proposed data augmentation methods brings accuracy improvements, we can provide a positive answer to RQ1. If none of the proposed data augmentation methods work, we cannot be sure of a negative answer to RQ1, i.e. there might be a data augmentation method that can bring accuracy improvements and we did not think of it. To minimize this risk, we propose a broad range of plausible data augmentation techniques.

% In order to answer our research questions, we conducted experiments with two deep learning \cite{LeCun-Nature-2015} models, a convolutional neural network (CNN) and a convolutional Long Short-Term Memory (ConvLSTM) network, on the HMOG data set \cite{Sitova-TIFS-2016} containing motion sensor signals from 100 mobile device users. We follow the experimental setting of Benegui et al. \cite{Benegui-Access-2020}, $(i)$ training the deep learning models on a subset of 50 users in a multi-way classification task and $(ii)$ employing the pre-trained models in a few-shot user identification task on the other 50 users. In the few-shot user identification task, only 20 data samples collected during user registration are available to train a Support Vector Machines (SVM) classifier. In order to train a binary SVM model for each user, 100 negative samples from other users are added to the training data. For a fair and realistic evaluation, the negative training samples and the negative test samples belong to disjoint sets of users, i.e. the SVM does not get to see data samples from actual attackers during training. 

In order to answer RQ1 and RQ2, we experiment with several data augmentation methods, namely adding random noise, temporal scaling, intensity scaling and warping, comparing the results with and without data augmentation. In order to answer RQ3, we consider to augment two independent deep learning models, namely a convolutional neural network (CNN) and a convolutional Long Short-Term Memory (ConvLSTM) network. %, as well as their combination. 
In order to answer RQ4, we propose to combine the data augmentation techniques that seem to bring accuracy improvements. We also test the combination that includes all data augmentation methods.

% During the registration process we collect a limited extent of identification samples through tap gesture performed on the screen, in the interest of not forcing the user undertake the same process a myriad number of times.

% Therefore, we propose the hypothesis in which the augmented HMOG dataset is further used to carry our experiments using our previously designed 6-layer CNN model, 6-layer ConvLSTM model \cite{benegui2019convolutional}, as well as a merge of feature vectors resulted from both models. Furthermore, we use a SVM classifier on top of the model extracted features to determine if the resulted dataset has an impact on user identification. 

% In summary, our contribution is threefold:
% \begin{itemize}
% \item We propose a novel approach to augment motion sensors yielded signals, in order to enhance user identification. 
% \item We propose to combine multiple augmented signals as well as experiment with the ratio between original and augmented number of samples.
% \item We conduct experiments on two benchmarks, showing if data augmentation is convenient for multiple types of models, or if only particular augmentation strategies yield better results.
% \end{itemize}

%We organize the rest of this paper as follows. We discuss related work in Section~\ref{sec_related_work}. We present the machine learning models and our data augmentation techniques in Section~\ref{sec_method}. We describe the experiments regarding the effectiveness of data augmentation methods in Section~\ref{sec_experiments}. Finally, we draw our conclusions in Section~\ref{sec_conclusion}.

\section{Related Work}
\label{sec_related_work}

Different studies explore the user identification task on mobile devices using various techniques. Among the first studies for biometric user identification, Vildjiounaite et al.~\cite{Vildjiounaite-ICPC-2006} used accelerometer-based gait recognition along with voice recognition to identify a user, the identification model being based on statistical features. Sitov\'a et al.~\cite{Sitova-TIFS-2016} approached the problem by analyzing human movement captured by different motion sensors, apprehending two specific sets of features: stability features and resistance features. During a tap gesture on the screen, motion data is collected and transformed into statistical features, which are given as input to a machine learning model. Recent research~\cite{Bo-IPCCC-2014,Buriro-CODASPY-2018,Buriro-ISBA-2017,Canfora-ICETE-2017,Ehatisham-JNCA-2018,Ku-Access-2019,Li-BIBM-2018,Neverova-Access-2016,Shen-Sensors-2016,Shi-WiMob-2011,Sitova-TIFS-2016,Sun-ECML-2017,Wang-Access-2019} shows that machine learning models generally attain better accuracy rates in the user identification process, compared to models based on statistical features~\cite{Vildjiounaite-ICPC-2006}. Among these machine learning models, a recent trend is to employ deep learning approaches \cite{Benegui-Access-2020,Neverova-Access-2016,Sun-ECML-2017}. Neverova et al.~\cite{Neverova-Access-2016} presented an approach based on recurrent neural networks by combining the two essential steps of machine learning (feature extraction and classification) into a single step. This is achieved through end-to-end learning. %Features are extracted passively, from biometric data, thus it does not require direct human action, the data being recorded in background. 
The method presented in~\cite{Neverova-Access-2016} requires a longer period of data gathering in order to produce optimal results. However, as noted by Benegui et al.~\cite{Benegui-Access-2020}, a user can also be identified from motion sensors by training a model on as few as 20 taps on the screen. This enables a fast registration and allows the coupling with explicit authentication systems based on face recognition or fingerprint scanning. % The motion sensor data is transformed into gray-scale images used as input for a convolutional convolutional neural network that produces embeddings further provided to a SVM binary classifier. Moreover, the method yields more robust embeddings and better accuracy over RNN based models such as the ones proposed by Neverova et al.~\cite{Neverova-Access-2016} and Sun et al.~\cite{Sun-ECML-2017}.
We hereby note that none of the methods mentioned so far, which are designed for user identification based on motion sensor data, study data augmentation. Nonetheless, we acknowledge that in related fields, e.g. human activity recognition, recent studies have shown that augmentation methods applied to time-series data can enhance classification results~\cite{Le-Guennec-AALTD-2016,Steven-Sensors-2018}. % Moreover, Guennec et al.~\cite{le2016data} demonstrated that time-series datasets can be better classified by CNN models using augmented samples. However,
To our knowledge, we are the first to study data augmentation on discrete motion sensor values used for user identification on mobile devices. 

\section{Methods}
\label{sec_method}

\subsection{Learning Models}
\label{sec_subsec}

In order to answer our research questions, we conducted experiments with two deep learning \cite{LeCun-Nature-2015} models, a CNN and a ConvLSTM, on the HMOG data set \cite{Sitova-TIFS-2016} containing motion sensor signals from 100 mobile device users. We follow the experimental setting of Benegui et al. \cite{Benegui-Access-2020}, $(i)$ training the deep learning models on a subset of 50 users in a multi-way classification task and $(ii)$ employing the pre-trained models in a few-shot user identification task on the other 50 users. As Benegui et al. \cite{Benegui-Access-2020}, we use a 6-layer CNN and a 6-layer ConvLSTM as pre-trained feature extractors, so we need to remove their Softmax layers. The last remaining fully-connected layer, which has 256 neurons, provides feature vectors (embeddings) of 256 components. A complete description of the CNN and the ConvLSTM architectures is given in \cite{Benegui-Access-2020}. We further utilize the embeddings resulting from either neural model as inputs to a Support Vector Machines (SVM) classifier, %We also consider the concatenation of the embeddings from both neural models into a single 512-dimensional feature vector. 
modeling the few-shot user identification task as a binary classification problem. In the few-shot user identification task, only 20 data samples collected during user registration are available to train the SVM classifier. 
In order to train a binary SVM model for each user, 100 negative samples from other users are added to the training data. For a fair and realistic evaluation, the negative training samples and the negative test samples belong to disjoint sets of users, i.e. the SVM does not get to see data samples from actual attackers during training. Benegui et al. \cite{Benegui-Access-2020} showed the benefits of modeling the user identification task as a binary classification problem with SVM instead of an outlier detection task with one-class SVM. %For the SVM, we employ two well-known kernel functions \cite{Taylor-CUP-2004}, namely the linear kernel and the Radial Basis Function (RBF) kernel. 

\subsection{Data Augmentation}
\label{sec_subsec}

Starting with the assumption that data augmentation can have a positive impact on the accuracy of the user identification system, we propose to experiment with different data augmentation techniques in order to assess their benefits.

% evaluate if the current approach yields better results in relation to the few-shot user identification system. Farther, we pursue the following augmentation strategies:
% \begin{itemize}
%     \item Random signal augmentation - Each signal value is multiplied with a random noise, thus generating a new sample
%     \item Scale-Crop augmentation - The signal is multiplied with a given factor $F$. If $F$ is greater than 1, then the signal is cropped to a well established signal length. Otherwise, if $F$ is smaller than 1, the scaled signal is zero-padded to fit a well established signal length.
%     \item Intensity augmentation - All the signal values are multiplied with the same factor 
%     \item Warp \cite{7797091}, from left to right (LTR) augmentation - For each signal, two points T1, T2 are randomly chosen, as follows: T1 is selected between the starting value of the signal and the middle value while T2 is selected between the middle value and the end of the signal. Further, values are stretched between T1, T2. Outliers remain unchanged.
%     \item Warp \cite{7797091}, from right to left (RTL) augmentation - Analog to left to right warp, in this case, the data is stretched from T2 to T1.
% \end{itemize}

% We observe in research literature that data augmentation can enhance the results of classification tasks \cite{DBLP:journals/corr/abs-1712-04621,Shorten-JBD-2019}. Thus, we propose to explore the following described augmentation methods.

\begin{figure}[!t]
\centering
\includegraphics[width=0.5\linewidth]{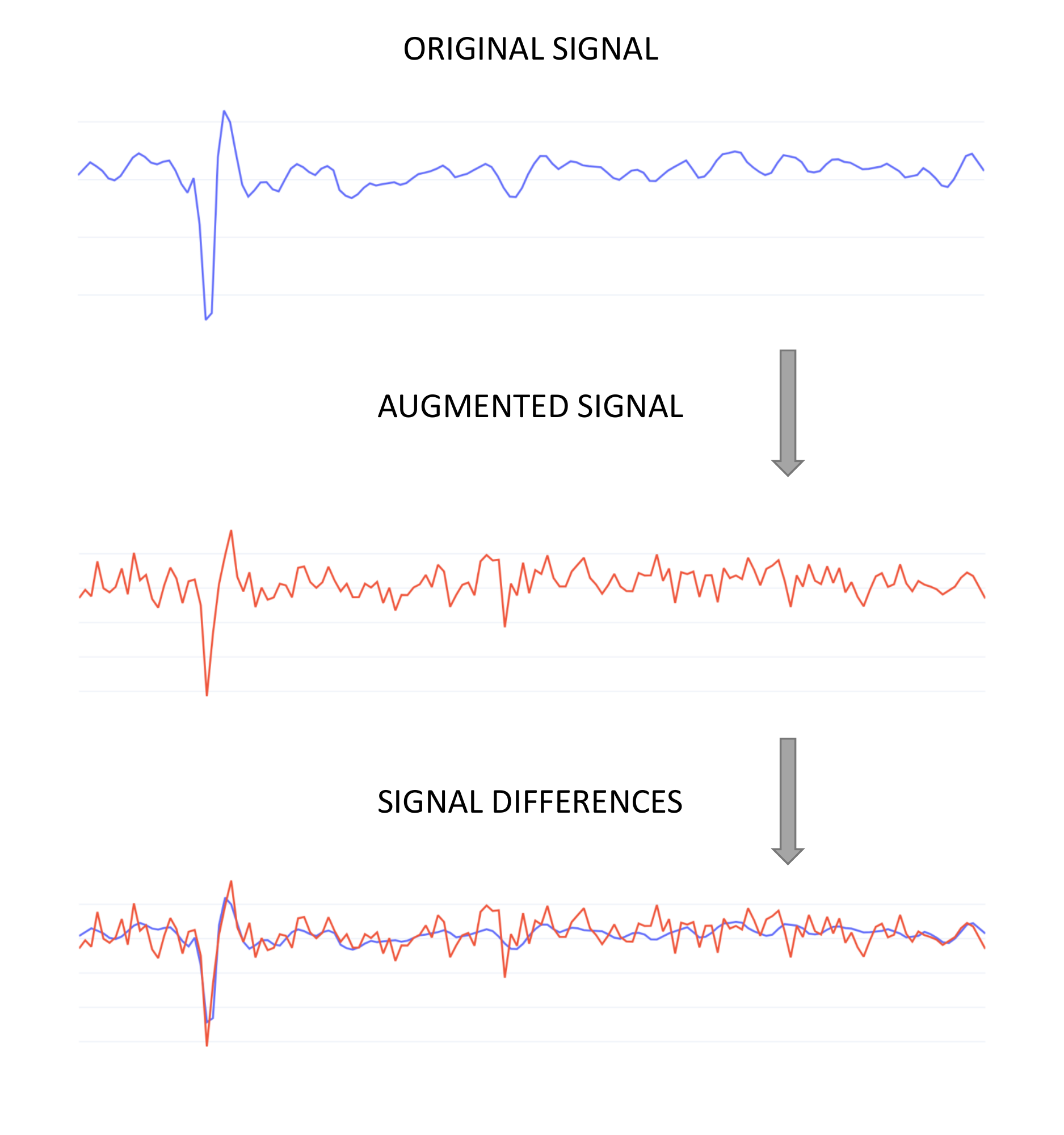}
\vspace{-0.6cm}
\caption{Augmentation of a signal with random Gaussian noise. Best viewed in color. }\label{fig_randn_aug}
\end{figure}

\noindent
{\bf Adding random noise.}
The first augmentation is to add a Gaussian noise signal $\vartheta$ that is randomly generated from a normal distribution $\mathcal{N}(\mu,\nu)$, where $\mu$ is the mean $\mu$ and $\sigma$ is the standard deviation. Given a motion sensor signal $S$ of length $|S| = n$, the addition of the Gaussian noise is formally expressed as follows:
\begin{equation}\label{eq_randn_aug}
    S_i = \vartheta_i + S_i, \; \vartheta_i \sim \mathcal{N}(\mu,\nu), \; \forall i \in \{1,...,n\}.
\end{equation}

We note that the amplitude of the noise signal and the degree to which it affects our signal $S$ are controlled through the parameter $\sigma$. In the experiments, we try out different values for $\sigma$. The effect of applying Eq.~(\ref{eq_randn_aug}) on a motion signal is illustrated in Figure~\ref{fig_randn_aug}.

\noindent
{\bf Temporal scaling.}
The second augmentation method scales the signal in the temporal domain based on a scaling factor $f_T$. When the scaling factor $f_T$ is greater than $1$, the length of the signal $S$ increases and the resulting signal is equally cropped on both sides to preserve the original signal length. When $f_T$ has a value lower than 1, the original signal gets contracted and the resulting signal is zero-padded at both ends in order to keep the initial signal length. In order to rescale the discrete signal, we apply linear interpolation \cite{Meijering-IEEE-2002}. % In Figure~\ref{fig_temp_scale_aug}, we illustrate the temporal scaling augmentation for a sub-unitary scale factor, i.e. for $f_T<1$.

% \begin{figure}[!t]
% \centering
% \includegraphics[width=0.75\linewidth]{figs/Signal augmentation - SCALE.pdf}
% \vspace{-0.6cm}
% \caption{Augmentation of a signal using temporal scaling with an over-unit scale factor. Best viewed in color.}\label{fig_scale_aug}
% \end{figure}

% \begin{figure}[!t]
% \centering
% \includegraphics[width=0.5\linewidth]{figs/Signal augmentation - CROP.pdf}
% \vspace{-0.3cm}
% \caption{Augmentation of a signal using temporal scaling with a sub-unitary scale factor. Best viewed in color.}\label{fig_temp_scale_aug}
% \end{figure}

\noindent
{\bf Signal intensity scaling.}
Given a signal $S$ and an intensity scale factor $f_I$, the augmented signal is obtained by multiplying each signal value $S_i$ with the scale factor. Formally, the intensity scaling augmentation of a signal $S$ is given by:
\begin{equation}\label{intensity_formula}
    S_i = f_I \cdot S_i, \; \forall i \in \{1,...,n\}.
\end{equation}
% Figure~\ref{fig_intensity_aug} illustrates the effect of the intensity scaling augmentation. 
In the experiments, we try out different values for $f_I$.

% \begin{figure}[!t]
% \centering
% \includegraphics[width=0.5\linewidth]{figs/Signal augmentation - INTENSITY.pdf}
% \vspace{-0.6cm}
% \caption{Augmentation of a signal using intensity scaling. Best viewed in color.}\label{fig_intensity_aug}
% \end{figure}

\begin{figure}[!t]
\centering
\includegraphics[width=0.5\linewidth]{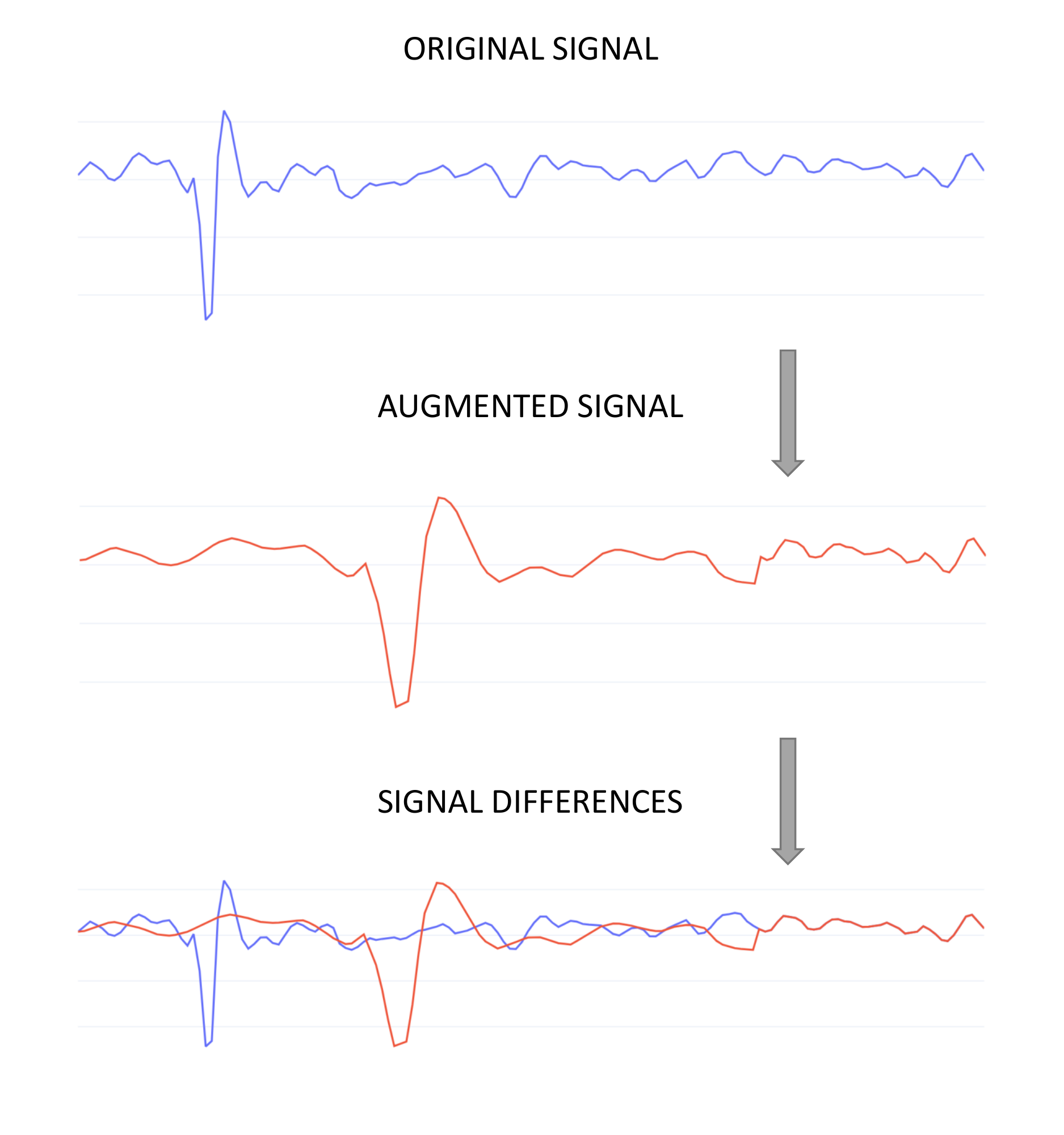}
\vspace{-0.6cm}
\caption{Augmentation of a signal using left-to-right warping. Best viewed in color.}\label{fig_ltr_aug}
\end{figure}

\noindent
{\bf Left-to-right warping.}
We propose an augmentation procedure in which the original signal is warped in the temporal domain by contracting the right side of the signal and expanding its left side. Given a motion sensor signal $S$ of length $|S|=n$, we first select two cutting points $t_1$ and $t_2$, randomly, as follows:
\begin{equation}\label{t1_formula}
t_1  \sim \mathcal{U}(\lfloor n/4 \rfloor, \lfloor n/2 \rfloor),\;
t_2  \sim \mathcal{U}(\lfloor n/2 \rfloor, \lfloor 3 \cdot n/4 \rfloor),
\end{equation}
where $\lfloor \cdot \rfloor$ is the flooring function and $\mathcal{U}(a, b)$ generates an integer value that is uniformly distributed between $a$ and $b$. The left part of the signal is stretched from $t_1$ to $t_2$. In the same time, the right part of the signal is contracted from $t_1$ to $t_2$. The discrete values are computed through linear interpolation. The resulting signal has the same length as the input signal. We illustrate the left-to-right warping of a signal in Figure~\ref{fig_ltr_aug}.

\noindent
{\bf Right-to-left warping.}
An analogous warping augmentation procedure is to contract the left side of the signal, while stretching its right side. We call this type of augmentation right-to-left warping. As for the left-to-right warping, we rely on the randomly-generated cutting points $t_1$ and $t_2$ to establish exactly how the signal is warped.

% \begin{figure}[!t]
% \centering
% \includegraphics[width=0.75\linewidth]{figs/Signal augmentation - WARP RTL.pdf}
% \caption{Visual representation of warp right to left signal augmentation. Best viewed in color.}\label{fig_rtl_aug}
% \end{figure}

\section{Experiments}
\label{sec_experiments}

\subsection{Data Set}\label{subsec_dataset}

We experiment on the HMOG \cite{Sitova-TIFS-2016} data set, which consists of discrete signals from mobile device motion sensors (gyroscope and accelerometer). Motion sensors yield values for three axes $(x,y,z)$ at roughly 100 Hz. We record values for 1.5 seconds during tap gestures on the screen, resulting in discrete signals of approximately 150 values. Signals are collected for 100 users, considering the first 200 tap events for each user. Hence, the resulting data set consists of 20,000 signal samples. Further, we divide the users in half, using the first half (50 users) to train the neural networks in a 50-way classification task and the second half for the few-shot user identification experiments. In the 50-way classification task, we employ an 80\%-20\% train-validation split, thus having 160 samples per user for training and 40 samples per user for validation. In the few-shot user identification experiments, we have 50 binary classification problems (one per user) in which the training set is composed of 20 positive and 100 negative samples and the test set is composed of another 100 positive and 100 negative samples. It is important to note that the 100 negative training samples are gathered from one subset of users and the 100 negative test samples are gathered from another (disjoint) subset of users. By adopting disjoint sets of users, we ensure that features representative for the attackers are not seen during training, resulting a in realistic scenario for our experiments.

\subsection{Experimental Setup}

\noindent
{\bf Evaluation metrics.}
%For the 50-way user classification experiments, we employ the classification accuracy. For the few-shot user identification experiments, 
We compute the accuracy, the false acceptance rate (FAR) and the false rejection rate (FRR) for each user. We then report the values averaged over the 50 users selected for the few-shot user identification experiments. %We note that the FAR is the ratio between the number of false acceptances (false positives) and the sum of all negative samples (false positives and true negatives). Similarly, the FRR is the ratio between the number of false rejections (false negatives) and the sum of all positives samples (false negatives and true positives).

\noindent
{\bf Parameter tuning for learning models.}\label{subsec_implementation}
% Pre-processing of the signals is required for the CNN model. Thus, we process each of the six axis signals to a fixed length of 150 values, using linear interpolation and apply De Bruijn's principle\cite{ralston1982bruijn} to obtain a unique arrangement of all axis. Further, the sequence is transformed into a gray-scale image with a fixed size of 25$\times$150 pixels, used as an input for our model. 
For the CNN model, we use the hyperparameters described in \cite{Benegui-Access-2020}, which provided optimal results on the validation set. We thus fix the learning rate to $10^{-3}$ and use mini-batches of 32 samples. The model is trained using Adam \cite{Kingma-ICLR-2015} for 50 epochs. To avoid overfitting, each fully-connected layer uses dropout at a rate of $0.4$. 
%The last fully connected layer is composed of 256 neurons and it is used for feature extraction in our experiments. This always results a 256-dimensional embeddings, handed as input for the binary classifier.
For the ConvLSTM architecture, we use the same hyperparameter settings as for the CNN model. Therefore, we set the learning rate to $10^{-3}$ and train the model for 50 epochs on mini-batches of 32 samples. Each fully-connected layer employs a dropout rate of 0.4. 
% Furthermore, the last fully connected layer has 256 neurons, yielding 256-dimensional embeddings. Moreover, ConvLSTM models use temporal data as input, thus, pre-processing applied to the motion sensors signals only consists of fixing their length to 150 values, using linear interpolation.
In the few-shot user identification task, we adopt binary SVM classifiers based on either a linear kernel or an RBF kernel \cite{Taylor-CUP-2004}. Throughout the experiments, we try out different values for the regularization parameter $C$ of the SVM, considering values in the set $\{1, 10, 100\}$. In order to compare the various SVM models in a balanced and fair setting, we automatically adjust the bias value of each SVM such that the difference between
the FAR and the FRR is less than 1\%.

\noindent
{\bf Baselines.}
We consider as baselines, the results attained by the SVM based on CNN or ConvLSTM embeddings, respectively, without data augmentation. %In the 50-way use classification experiment, the CNN model provides a training accuracy of $94.17\%$ and a validation accuracy of $90.75\%$. In the few-shot used identification experiment, the SVM based on the pre-trained CNN embeddings attains an average test accuracy rate of $96.37\%$ with an FAR of $3.30\%$ and an FRR of $3.96\%$. Similarly, by applying the SVM on the embeddings provided by the pre-trained ConvLSTM, we attain an average accuracy of $96.18\%$ with an FAR of $4.00\%$ and an FRR of $3.64\%$.

% using 1 as the value for the regularization parameter C and applying RBF as the kernel of choice. 
% for the same kernel RBF and a C value of 1.

\noindent
{\bf Data augmentation scenarios.}
We note that the degree of augmentation is not only reflected by the hyperparameter choices for the data augmentation methods, but also by the number of augmented samples. Therefore, in our experiments, we explore two different ratios between the number of original samples and the number of augmented samples, as follows. In the first augmentation scenario, we employ an augmentation ratio of $1\times$, so that during the training phase, each original data sample is copied and augmented once. This results in a training set with 40 positive samples (20 original and 20 augmented) and 200 negative samples (100 original and 100 augmented). In the second augmentation scenario, we employ an augmentation ratio of $0.5\times$, so that during the training phase, one in every two original data samples is copied and augmented once. This results in a training set with 30 positive samples (20 original and 10 augmented) and 150 negative samples (100 original and 50 augmented). %In the third augmentation scenario, we employ an augmentation ratio of $1\times$ only for the positive data samples. This results in a training set with 40 positive samples (20 original and 20 augmented) and 100 original negative samples. This augmentation scenario should show the benefit of using data augmentation to rebalance the training set, at least to some degree. 
We hereby note that we do not use data augmentation during the testing phase, i.e. we keep the same number of test samples, 100 positive and 100 negative per user. 

\noindent
{\bf Parameter tuning for data augmentation methods.}\label{subsec_parameter}
% SVM classifier is applied for each user of the 50 users in our data% set. During the training phase, we account for 20 positive samples and 100 negative samples while in the testing phase, we use a different subset of 100 positive and 100 negative samples.
We tune the parameters of each augmentation method in order to assess which configuration provides the highest improvements in terms of identification accuracy.
We carry out the augmentation based on random Gaussian noise using different values for the standard deviation value $\sigma$, considering $\sigma \in $ \{0.0125, 0.025, 0.05, 0.1, 0.2, 0.3, 0.4, 0.5\}. We note that the amplitude of the Gaussian noise is directly proportional to the value of $\sigma$, so greater values result in larger deviations from the original signal. 
For temporal scaling, the degree to which a signal is stretched or contracted is controlled by the parameter $f_T$, which represents the temporal scaling factor. %Through linear interpolation, the signal is either stretched and cropped or compressed and zero-padded, respecting the initial signal length. 
In our experiments, we select $f_T$ within a range of values that results in either stretching (when $f_T > 1$) or contracting (when $f_T < 1$) the original signals. For $f_T$, we considered values in the set \{0.8, 0.9, 0.95, 0.975, 0.9875, 1.0125, 1.025, 1.05, 1.1, 1.2\}. % While $F=1$ is being a neutral value that does not affect the signal for this type of augmentation, we observe an accuracy change of statistical significance with slim adjustments to this value.
For intensity scaling, the degree to which the amplitude of a signal is exaggerated or flattened is controlled by the parameter $f_I$, which represents the intensity scaling factor. For $f_I$, we considered values in the set \{0.8, 0.9, 0.95, 0.975, 0.9875, 1.0125, 1.025, 1.05, 1.1, 1.2\}.
For signal warping, we consider the direction of the warp, left-to-right ($L\!\rightarrow\!R$) or right-to-left ($L\!\leftarrow\!R$), as the only parameter that requires tuning.
In the subsequent experiments, we report accuracy rates only for the optimal parameter values, specifying in each case the corresponding hyperparameter value. The parameters are validated by fixing the data representation to the embeddings provided by the CNN. We then use the same parameters for the ConvLSTM, in order to avoid overfitting in hyperparameter space.

\begin{table*}[!t]
\centering
\caption{\label{cnn_lstm_augmentation_table_1} Results for the few-short user identification task with various SVM classifiers trained on embeddings provided by pre-trained CNN or ConvLSTM models, with and without data augmentation. Each augmentation procedure is evaluated in two augmentation scenarios. In each case, results are reported only for the optimal hyperparameter values. Accuracy, FAR and FRR scores represent the average values computed on 50 users. Results that exceed the baseline accuracy rates are marked with asterisk.}
\setlength\tabcolsep{1.2pt}
\scriptsize{
\begin{tabular}{|l|c|l|c|c|c|c|l|c|c|c|c|}
\hline
\multicolumn{1}{|c|}{{Augmentation}} &
  \multicolumn{1}{c|}{Parameter} &
  \multicolumn{5}{c|}{SVM+CNN embeddings} &
  \multicolumn{5}{c|}{SVM+ConvLSTM embeddings} \\ \cline{3-12} 
\multicolumn{1}{|c|}{method} & \multicolumn{1}{c|}{value} & Kernel & C   & Accuracy     & FAR    & FRR                 & Kernel & C   & Accuracy     & FAR     & FRR \\ 
\hline
\hline
\multicolumn{12}{|l|}{\textbf{No augmentation}} \\
\hline
-                 & -                           & RBF   & 1 & $96.37\%$ & $3.30\%$ & $3.96\%$               & RBF & 1   & $96.18\%$   & $4.00\%$  & $3.64\%$  \\
\hline
\multicolumn{12}{|l|}{\textbf{Augmentation of all samples with ratio $\mathbf{1\times}$}} \\
\hline
Random noise     & $\sigma=0.025$        & Linear & 100 & $96.54\%^\star$ & $3.46\%$ & $3.45\%$      & Linear & 1   & $95.63\%$   & $4.30\%$  & $4.44\%$  \\
\hline
Temporal scaling        & $f_T=0.975$           & Linear & 100 & $96.48\%^\star$ & $4.01\%$ & $3.83\%$      & Linear & 10  & $96.77\%^\star$ & $3.30\%$  & $3.15\%$  \\
\hline
Intensity scaling       & $f_I=0.95$            & Linear & 1   & $96.50\%^\star$ & $3.58\%$ & $3.41\%$      & Linear & 1   & $93.63\%$ & $6.60\%$  & $6.14\%$ \\
\hline
Warping                 & $L\!\leftarrow\!R$    & RBF    & 1   & $94.87\%$ & $4.96\%$ & $5.29\%$            & Linear & 1   & $95.94\%$ & $3.98\%$  & $4.14\%$ \\
\hline
\multicolumn{12}{|l|}{\textbf{Augmentation of all samples with ratio $\mathbf{0.5\times}$}} \\
\hline
Random noise     & $\sigma=0.05$         & Linear & 1   & $96.54\%^\star$ & $3.42\%$ & $3.49\%$      & Linear & 10  & $96.48\%^\star$ & $3.96\%$  & $3.07\%$ \\
\hline
Temporal scaling        & $f_T=1.05$            & Linear & 100 & $96.77\%^\star$ & $3.26\%$ & $3.19\%$      & Linear & 1   & $94.89\%$ & $5.08\%$  & $5.13\%$ \\
\hline
Intensity scaling       & $f_I=1.0125$          & Linear & 1   & $96.41\%^\star$ & $3.42\%$ & $3.76\%$      & Linear & 10  & $95.35\%$ & $4.74\%$  & $4.57\%$  \\
\hline
Warping                 & $L\!\rightarrow\!R$   & RBF    & 1   & $94.49\%$ & $5.24\%$ & $5.79\%$            & Linear & 10  & $95.56\%$ & $4.36\%$  & $4.53\%$  \\
% \hline
% \multicolumn{12}{|l|}{\textbf{Augmentation of positive samples with ratio $\mathbf{1\times}$}} \\
% \hline
% Random noise     & $\sigma=0.05$         & Linear & 1   & $96.34\%$ & $3.66\%$ & $3.66\%$            & RBF    & 10 & $92.64\%$ & $7.38\%$ & $7.33\%$ \\
% \hline
% Temporal scaling        & $f_T=1.025$           & Linear & 1   & $96.63\%^\star$ & $5.76\%$ & $5.97\%$      & RBF    & 10 & $90.76\%$ & $9.28\%$ & $9.19\%$ \\
% \hline
% Intensity scaling       & $f_I=0.95$            & Linear & 100 & $96.12\%$ & $4.06\%$ & $3.70\%$            & RBF    & 10 & $91.45\%$ & $8.54\%$ & $8.57\%$ \\
% \hline
% Warping                 & $L\!\leftarrow\!R$    & RBF    & 10  & $94.70\%$ & $5.56\%$ & $5.03\%$            & RBF    & 100 & $91.10\%$ & $8.86\%$ & $8.95\%$ \\
\hline
\end{tabular}}
%}
\end{table*}

\subsection{Results with Independent Augmentations}\label{subsec_results}

%With respect to all the possible values for the augmentation methods parameters, we conduct experiments in order do determine which value produces the best results. In consequence, empirical results in this section are based on the following best performing parameters: Random, $\sigma$ equal to 0.025; Intensity, using $I_f$ = 0.95; Scale-Crop, with $F$=0.975 and finally, Warp using RTL direction.
% \subsubsection{Augmentation results}\label{full_augmentation}

In Table~\ref{cnn_lstm_augmentation_table_1}, we present the empirical results obtained by various SVM classifiers based on CNN or ConvLSTM features for different augmentation scenarios. For each type of augmentation, we include the scores attained only for the best performing parameters. %Correspondingly, we take advantage of the best performing parameters on the binary classifier based on CNN features and implement them for the LSTM, with the purpose to identify if the augmentation produces performance improvements for both models. 

\noindent
{\bf Augmentation of all samples with ratio $\mathbf{1\times}$.}
When we copy and augment all training samples exactly once, we observe that the SVM based on CNN embeddings performs better than the baseline SVM for three independent augmentation techniques: random noise addition, temporal scaling and intensity scaling. However, the differences between the baseline SVM based on CNN embeddings and the SVM based on CNN embeddings with data augmentation are slim, the maximum improvement being $+0.17\%$. The random noise and the temporal scaling augmentation methods yield their best accuracy rates using  an SVM based on a linear kernel and a regularization of $C=100$. The intensity scaling augmentation works better with an SVM based on a linear kernel with $C=1$. With respect to the SVM based on ConvLSTM embeddings, we observe accuracy improvements ($+0.59\%$) only when the data is augmented through temporal scaling. We notice that none of the observed improvements are statistically significant. We also note that warping is the only augmentation technique that seems to degrade performance for both CNN and ConvLSTM embeddings.

\noindent
{\bf Augmentation of all samples with ratio $\mathbf{0.5\times}$.}
If the number of augmented samples was too high in the first augmentation scenario, we should be able to observe this problem in the second augmentation scenario, in which the augmentation ratio is $0.5\times$. Considering the comparative results presented in Table~\ref{cnn_lstm_augmentation_table_1}, we notice moderate changes in terms of accuracy rates. As in the first scenario, the same three data augmentation methods bring performance improvements over the baseline SVM based on CNN embeddings. Temporal scaling generates an accuracy improvement of $+0.40\%$ for the linear kernel and $C=100$, becoming the best augmentation method, followed by the random noise augmentation with an increase of $+0.17\%$ (just as in the first augmentation scenario). With respect to the SVM based on ConvLSTM embeddings, we observe that the baseline is surpassed only when the data is augmented with random noise. Considering that  we attained better results with temporal scaling for the augmentation ratio $1\times$, we conclude that the results reported for the SVM based on ConvLSTM embeddings are inconsistent.

% \begin{table}[!t]
% \centering
% \caption{Results for the few-short user identification task with an SVM classifier trained on concatenated CNN and ConvLSTM embeddings, with and without data augmentation. In each case, results are reported only for the optimal hyperparameter values. Accuracy, FAR and FRR scores represent the average values on 50 users.}
% \label{experiments_table_3}
% \scriptsize{%
% %\setlength\tabcolsep{2.5pt}

% \begin{tabular}{|l|c|l|c|c|c|c|}
% \hline
% \multicolumn{1}{|c|}{Augmentation}    & Parameter     & \multicolumn{5}{c|}{CNN+ConvLSTM embeddings} \\
% \cline{3-7} 
% \multicolumn{1}{|c|}{method}          & value         & Kernel & C & Accuracy     & FAR    & FRR    \\
% \hline
% \hline
% \multicolumn{7}{|l|}{\textbf{No augmentation}} \\ \hline
% -                   & -                     & RBF    & 100 & $96.74\%$ & $3.10\%$ & $3.41\%$ \\
% \hline
% \hline
% \multicolumn{7}{|l|}{\textbf{Augmentation of all samples with ratio $\mathbf{1\times}$}} \\
% \hline
% Random noise        & $\sigma=0.025$        & RBF    & 1 & $96.69\%$ & $3.16\%$ & $3.45\%$ \\
% \hline
% Temporal scaling    & $f_T=0.975$           & Linear & 1 & $96.67\%$ & $3.44\%$ & $3.21\%$ \\
% \hline
% Intensity scaling   & $f_I=0.95$            & Linear & 1 & $96.40\%$ & $3.68\%$ & $3.52\%$ \\
% \hline
% Warping             & $L\!\leftarrow\!R$    & RBF    & 1 & $94.92\%$ & $5.20\%$ & $4.95\%$ \\
% \hline
% \end{tabular}
% }
% \end{table}

% \usepackage{multirow}
\begin{table*}[!t]
\centering
\caption{Results for the few-short user identification task with various SVM classifiers trained on CNN or ConvLSTM embeddings, with and without aggregated data augmentation methods. Aggregated augmentations are evaluated in one augmentation scenario. Results are reported only for the optimal hyperparameter values. Accuracy, FAR and FRR scores represent the average values computed on 50 users. Results that exceed the baseline accuracy rates are marked with asterisk.}
\label{experiments_table_2}
%\scalebox{0.9}{%
%\setlength\tabcolsep{0.5pt}
\scriptsize{
\begin{tabular}{|l|l|c|c|c|c|l|c|c|c|c|}
\hline
\multicolumn{1}{|c|}{\multirow{2}{*}{Augmentation method}} & \multicolumn{5}{c|}{SVM+CNN embeddings} & \multicolumn{5}{c|}{SVM+ConvLSTM embeddings} \\ \cline{2-11} 
\multicolumn{1}{|c|}{}        & Kernel & C   & Acc.     & FAR    & FRR    & Kernel & C  & Acc.     & FAR    & FRR    \\ 
\hline
\hline
\multicolumn{11}{|l|}{\textbf{No augmentation}} \\
\hline
-                                   & RBF   & 1 & $96.37\%$ & $3.30\%$ & $3.96\%$               & RBF & 1   & $96.18\%$   & $4.00\%$  & $3.64\%$  \\
\hline

\multicolumn{11}{|l|}{\textbf{Augmentation of all samples with ratio $\mathbf{1\times}$}}                                            \\
\hline
All augmentation methods            & Linear & 1   & $96.29\%$ & $3.60\%$ & $3.82\%$            & RBF    & 1  & $93.76\%$ & $6.18\%$ & $6.30\%$ \\
\hline
%All augmentations combined \textsuperscript{2} & Linear & 10  & 95.63\% & 4.48\% & 4.26\% & RBF    & 1  & 93.27\% & 6.60\% & 6.87\% \\ \hline
Random noise+temporal scaling    & Linear & 100 & $96.48\%^\star$ & $3.82\%$ & $3.21\%$      & RBF    & 1  & $93.73\% $& $6.32\%$ & $6.22\%$ \\
%\hline
% \multicolumn{11}{|l|}{\textbf{Augmentation of positive samples with ratio $\mathbf{1\times}$}}  \\
% \hline
% All augmentation methods            & Linear & 100 & $96.36\%$ & $3.60\%$ & $3.68\%$            & RBF    & 1  & $93.33\%$ & $6.68\%$ & $6.67\%$ \\
% \hline
% %All augmentations combined \textsuperscript{2}      & Linear & 100 & 96.12\% & 3.88\% & 3.88\% & RBF    & 1  & 92.38\% & 7.52\% & 7.72\% \\ \hline
% Random noise + temporal scaling    & Linear & 1   & $96.33\%$ & $3.58\%$ & $3.76\%$            & RBF    & 10 & $92.54\%$ & $7.56\%$ & $7.35\%$ \\
\hline
\end{tabular}}
%}
\end{table*}

\subsection{Results with Combined Augmentations}\label{subsec_results}

% Empirical results for the user identification task performed by a SVM classifier using different aggregations of augmented feature samples. In the selected random augmentation \textsuperscript{1} subset we randomly choose augmented samples from all augmentation types. Our second method employes a subset with all augmentations combined \textsuperscript{2}, such that for each number of original samples we add to the subset the same number of samples from each augmentation. Lastly, our 3rd method adds an equal number of samples from each of first two best performing augmentations, namely Random and Intensity\textsuperscript{3}. All of the augmented samples are selected considering the best performing augmentation parameter. Accuracy, FAR and FRR represent the average values obtained from the 50 users subsample involved in the identification problem. The results that exceed the initial accuracy of the CNN model, 96.37\%, are represented using the $\odot$ symbol. For each best performing augmentation parameter, we outline the SVM kernel used and the regularization parameter C value.

After experimenting with various data augmentation methods and learning how they impact performance, we explore the augmentation with combined methods, which may lead to further performance boosts. We consider two alternative mixtures of data augmentation methods. On the one hand, we consider the combination of all our data augmentation methods. On the other hand, we consider the combination of the best two methods, namely random noise and temporal scaling. We present the corresponding results for the first augmentation scenario, in which the ratio is $1\times$, in Table~\ref{experiments_table_2}.
% outlines empirical results obtained by aggregating different augmentation methods. We aim to validate if different variations of the augmented signals improve the performance of the classifier or makes it worse. 
%\subsubsection{Augmentation of all samples with ratio $1\times$}
When the SVM is based on ConvLSTM features, it seems that neither combination of data augmentation methods is able to surpass the baseline. When the SVM is based on CNN embeddings, the sole combination that slightly outperforms the baseline results is composed of random noise and temporal scaling. The $+0.11\%$ accuracy improvement is obtained using an SVM based on a linear kernel with $C=100$. We hereby note that aggregating different augmentations does not contribute to significant improvements.

% \subsubsection{Augmentation of positive samples with ratio $1\times$}

% When we augment only the positive data samples, we do not observe any improvements. Moreover, it appears that data augmentation has a significant negative impact on the SVM based on ConvLSTM embeddings (accuracy drops are around $3\%$).

\section{Conclusion}
\label{sec_conclusion}

In this paper, we have studied different augmentation strategies for signals generated by motion sensors, with the intention of answering a set of research questions regarding the usefulness of data augmentation for the few-shot user identification problem. We performed a set of experiments with various data augmentation approaches using two state-of-the-art neural architectures, a CNN and a ConvLSTM, allowing us to answer the proposed research questions. We conclude our work by answering our research questions below:
\begin{itemize}
    \item RQ1: Can data augmentation bring accuracy improvements? \\
    Answer: In order to answer this question, we tried out multiple augmentation methods such as adding random noise, temporal scaling, intensity scaling and warping. We observed performance improvements (under $0.6\%$) for all methods, besides warping (see Table~\ref{cnn_lstm_augmentation_table_1}). In summary, the answer to RQ1 is affirmative, although the improvements are not statistically significant. 
    \item RQ2: Which of the proposed data augmentation methods brings accuracy improvements? \\
    Answer: Among the considered augmentation methods, we discovered that adding random noise, temporal scaling and intensity scaling can bring performance improvements. However, these improvements are not consistent across machine learning models and augmentation scenarios (see Table~\ref{cnn_lstm_augmentation_table_1}).
    \item RQ3: Are the data augmentation methods generic or specific to certain machine learning models? \\
    Answer: We considered to augment the data for two models, one based on CNN embeddings and one based on ConvLSTM embeddings. In most cases, we observed performance gains for the SVM model based on CNN embeddings (see Table~\ref{cnn_lstm_augmentation_table_1}). In very few cases, we noticed improvements for the SVM based on ConvLSTM embeddings (see Table~\ref{cnn_lstm_augmentation_table_1}). %When the embeddings are concatenated, data augmentation does not seem to help at all (see Table~\ref{experiments_table_3}). 
    We thus conclude that the data augmentation methods do not generalize across different models.
    \item RQ4: Can mixtures of data augmentation techniques bring accuracy improvements? \\
    Answer: We conducted experiments by aggregating the best two augmentation methods, as well as by aggregating all the data augmentation methods. Aggregating multiple augmentation methods does not seem to be effective (see Table~\ref{experiments_table_2}), so the answer to RQ4 is negative.
\end{itemize}

Looking at the overall picture, we conclude that data augmentation is not useful for few-shot user identification based on motion sensor data. We also notice that the augmentation hyperparameters ($\sigma$, $f_T$ and $f_I$) that provided the best results tend to correspond to the smallest changes on the original signals. This indicates that data augmentation is rather harmful, distorting or covering the patterns useful for discriminating registered users from attackers. In this context, we do not recommend data augmentation on discrete signals recorded by motion sensors. In future work, we aim to explain~\cite{Barbalau-ECML-2020} why the plain models (without augmentation) obtain such good results. Our intuition is that the models rely on features that are sensitive to changes brought by data augmentation. %To confirm it, we will look into visualization methods, e.g.~Grad-CAM [A], or explainable AI methods, e.g.~[B].

\section*{Acknowledgment}
The research leading to these results has received funding from the EEA Grants 2014-2021, under Project contract no. EEA-RO-NO-2018-0496.


\bibliography{references}{} 
\bibliographystyle{splncs04}

\end{document}